%% file: main.tex
\documentclass[10pt,twocolumn,letterpaper]{article}

\usepackage{cvpr}              
\usepackage{booktabs}
\usepackage{multirow}

\usepackage[accsupp]{axessibility}  

\input{preamble}
\definecolor{cvprblue}{rgb}{0.21,0.49,0.74}
\usepackage[pagebackref,breaklinks,colorlinks,allcolors=cvprblue]{hyperref}

\def\paperID{43460} 
\def\confName{CVPR}
\def\confYear{2026}

\title{
\textit{CoIn3D}: Revisiting Configuration-Invariant Multi-Camera 3D Object Detection}

\author{
Zhaonian Kuang$^{1,2}$ $^*$ \quad
Rui Ding$^{1}$  \quad
Haotian Wang$^{2}$ \quad
Xinhu Zheng$^{2}$ $^\dagger$ \quad 
Meng Yang$^{1}$ $^\dagger$ \quad
Gang Hua$^{3}$
\\[2mm]
$^1$ National Key Laboratory of Human-Machine Hybrid Augmented Intelligence,\\ Institute of Artificial Intelligence and Robotics, Xi'an Jiaotong University\\
$^2$ Intelligent Transportation Thrust of the Systems Hub, HKUST(GZ)\quad
$^3$ Amazon Alexa AI\quad
}

\begin{document}
\maketitle

{\let\thefootnote \relax
\footnote{
\hangindent=1.8em
Codes are available \href{https://github.com/kwong292521/CoIn3D}{\nolinkurl{here}}.\\
$^*$ The author was funded while visiting HKUST(GZ).\\
$^\dagger$ Corresponding Authors.
}}

\input{sec/0_abstract}    
\input{sec/1_intro}
\input{sec/2_relate_work}

\input{sec/3_revisit}

\input{sec/4_methodology}
\input{sec/5_experiments}

\input{sec/6_conclusion}
\input{sec/7_ack}

{
    \small
    \bibliographystyle{ieeenat_fullname}
    \bibliography{main}
}

\input{sec/X_suppl}

\end{document}

%% file: sec/0_abstract.tex
\begin{abstract}
Multi-camera 3D object detection (MC3D) has attracted increasing attention with the growing deployment of multi-sensor physical agents, such as robots and autonomous vehicles.
However, MC3D models still struggle to generalize to unseen platforms with new multi-camera configurations.
Current solutions simply employ a meta-camera for unified representation but lack comprehensive consideration.  
In this paper, we revisit this issue and identify that the devil lies in spatial prior discrepancies across source and target configurations, including different intrinsics, extrinsics, and array layouts. 
To address this, we propose \textbf{CoIn3D}, a generalizable MC3D framework that enables strong transferability from source configurations to unseen target ones. CoIn3D explicitly incorporates all identified spatial priors into both feature embedding and image observation through \textbf{spatial-aware feature modulation (SFM)} and \textbf{camera-aware data augmentation (CDA)}, respectively. SFM enriches feature space by integrating four spatial representations, such as focal length, ground depth, ground gradient, and Plücker coordinate. CDA improves observation diversity under various configurations via a training-free dynamic novel-view image synthesis scheme. Extensive experiments demonstrate that CoIn3D achieves strong cross-configuration performance on landmark datasets such as NuScenes, Waymo, and Lyft, under three dominant MC3D paradigms represented by BEVDepth, BEVFormer, and PETR.
\end{abstract}

%% file: sec/1_intro.tex
\section{Introduction}

\label{sec:intro}

Driven by the demand for reliable target object localization, multi-camera 3D object detection (MC3D) \cite{wang2022detr3d,liu2022petr,li2023bevdepth,huang2022bevdet4d,li2024bevformer,yang2023bevformer,liu2023sparsebev} has gained increasing attention in recent years. MC3D has been widely adopted in physical agents such as autonomous vehicles \cite{li2023delving,chen2024end} and robots \cite{chen2024survey}.  Compared with explicit LiDAR-based counterparts \cite{yin2021center,lang2019pointpillars,jin2025unimamba,fan2023super,liu2022bevfusion,li2024fully}, vision-based methods rely on multi-camera configurations, including intrinsics, extrinsics, and array layouts, to implicitly perceive spatial structures from image observations.

\begin{figure}[t]
  \centering
   \includegraphics[width=1.0\linewidth]{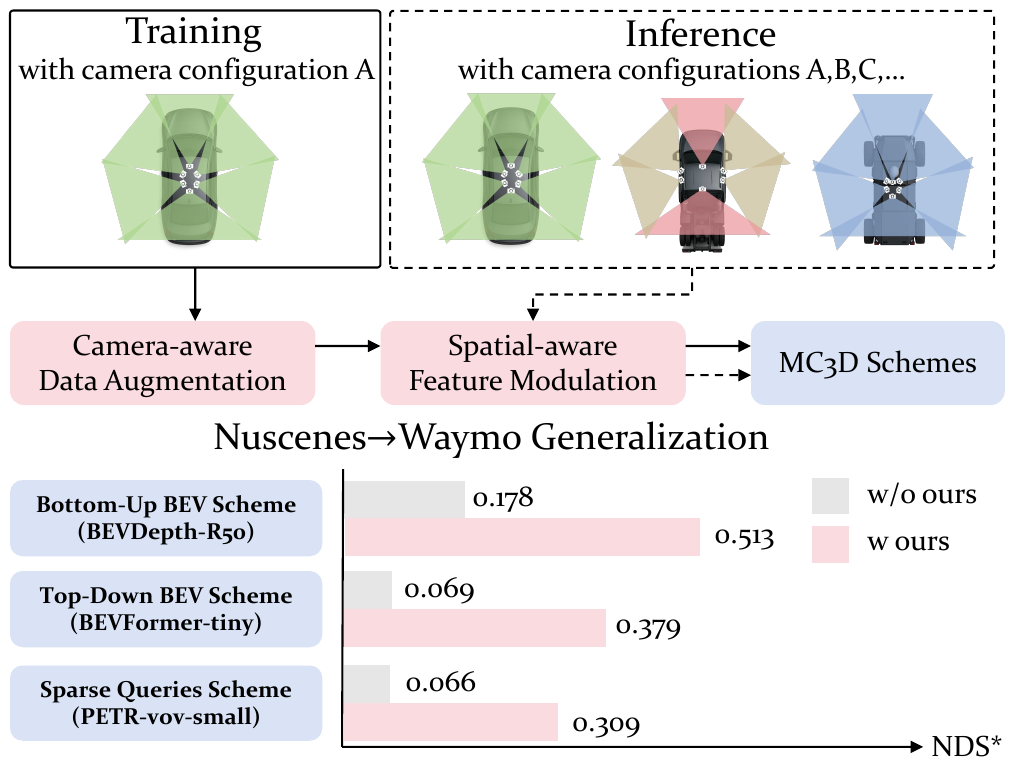}
   \caption{CoIn3D effectively enables model transferability from source configuration A to unseen target configurations B, C, ..., covering variations in intrinsics, extrinsics, and array layouts. Our framework can be applied to three dominant MC3D paradigms, represented by BEVDepth \cite{li2023bevdepth}, BEVFormer \cite{li2024bevformer}, and PETR \cite{liu2022petr}.}
   \label{fig:top-figure}
\end{figure}

Although current MC3D methods employ camera configurations to help connect image observations and spatial perception, their focus primarily lies in seeking vision invariance for unified image representation rather than achieving camera invariance for cross-configuration applications. In this paper, we present that the latter also plays a crucial role in MC3D model generalization.

Specifically, the configuration gap makes MC3D models perform poorly when tested on unseen configurations, leading to inflexibility and high costs in real-world deployment. A platform with a new configuration requires recollecting and reannotating data to retrain a specialized model.

To tackle the generalization problem caused by the configuration gap, some previous works \cite{yang2024geometry,zheng2023cross,li2024unidrive} directly warp images into meta-camera. Specifically, they resize and crop the image to align focal length and field-of-view (FoV), and then warp image based on spherical \cite{yang2024geometry} or cylindrical \cite{li2024unidrive} 3D space assumption to align extrinsics and arrays. However, warping input image causes resolution reduction and 3D scene structure distortion.

Some other works \cite{wang2023towards,lu2025towards,chang2024unified} treat all cameras as possessing a virtual meta-focal and rescale the predicted depth during training and testing. PD-BEV \cite{lu2025towards} and UDGA-BEV \cite{chang2024unified} further constrain feature and depth consistency for better generalization. In addition, \cite{kumar2025charm3r} introduces object bottom ground depth to address camera height gaps. However, these methods do not fully and explicitly consider camera configurations, and they cannot be applied to all MC3D paradigms due to their depth-based design.

This paper comprehensively revisits the impact of configurations on MC3D generalization. We identify that the devil lies in the spatial prior discrepancies across source and target configurations. First, regarding intrinsics, different focal lengths cause object pixel size ambiguity and different FoVs lead to different scene perspective. Second, regarding extrinsic, different camera mounting positions and orientations alter ground plane geometries and observed scene geometries. Both of these two factors affect model's spatial understanding. Lastly, different array layouts cause different camera numbers and overlap region, which affect the pattern of multi-camera correlation and feature fusion.

To address this, we propose CoIn3D, a generalizable MC3D framework that enables effective transferability from source configurations to unseen target configurations. CoIn3D explicitly incorporates identified spatial priors into both feature embedding and image observation through spatial-aware feature modulation (SFM) and camera-aware data augmentation (CDA), respectively.

Our SFM enriches the feature space by explicitly integrating four spatial representations. Specifically, we introduce four over-complete pixel-level spatial embeddings that formulate camera configurations. Among them, the inverse focal map formulates focal length, the ground depth and gradient maps jointly formulate ground geometry, and the Plücker raymap provides a holistic representation of camera configurations. 

We first obtain focal-invariant features by multiplying the inverse focal map with image features. Then, we concatenate the ground depth, gradient, and Plücker raymap into a mixture prior map. The mixture prior map is encoded into a spatial feature embedding and added to the focal-invariant feature to obtain spatial-embedded feature. We further concatenate four raw prior maps into the spatial-embedded feature to obtain spatial-aware feature. This spatial-aware feature can be used by all MC3D schemes.

Our CDA introduces a cost-efficient, training-free novel-view image synthesis scheme based on 3D Gaussian splatting (3DGS) to further enhance the generalization of MC3D. Specifically, we reconstruct ego-centric texture point clouds for each annotated data frame and transform them into a 3D Gaussian representation using predefined parameters. By randomly sampling cameras with diverse configurations, these Gaussians can be dynamically rendered into novel-view images for training.

Our framework can be widely applied to all dominant MC3D paradigms, including bottom-up BEV, top-down BEV, and sparse queries schemes, for which we choose three representative base models: BEVDepth \cite{li2023bevdepth}, BEVFormer \cite{li2024bevformer}, and PETR \cite{liu2022petr}, respectively.

Our framework is evaluated across three real-world landmark datasets with different camera configurations, including Nuscenes \cite{caesar2020nuscenes}, Waymo \cite{sun2020scalability}, and Lyft \cite{3d-object-detection-for-autonomous-vehicles}. Experiments show that our scheme achieves significant gains on all paradigms under different cross-dataset settings as shown in Fig. \ref{fig:top-figure}. In addition, our scheme achieves state-of-the-art (SOTA) performance in all settings based on BEVDepth.

The main contributions can be summarized as follows:
\begin{itemize}
    \item We revisit the influence of multi-camera configurations, and identify that the devil lies in spatial prior discrepancies across source and target configurations.
    \item We propose SFM to enrich feature by explicitly integrating four spatial representations, including focal length, ground depth, ground gradient, and Plücker coordinate.
    \item We propose CDA, a cost-efficient, training-free novel-view image synthesis scheme based on 3D Gaussian splatting, to dynamically render augmented training images with diverse configurations.
    \item Our framework consistently achieves significant gains on all MC3D paradigms under different cross-dataset settings and achieves SOTA based on BEVDepth. 
\end{itemize}

%% file: sec/2_relate_work.tex
\section{Related Work}
\label{sec:relate_works}
\begin{figure*}
  \centering
  \includegraphics[width=1.0\linewidth]{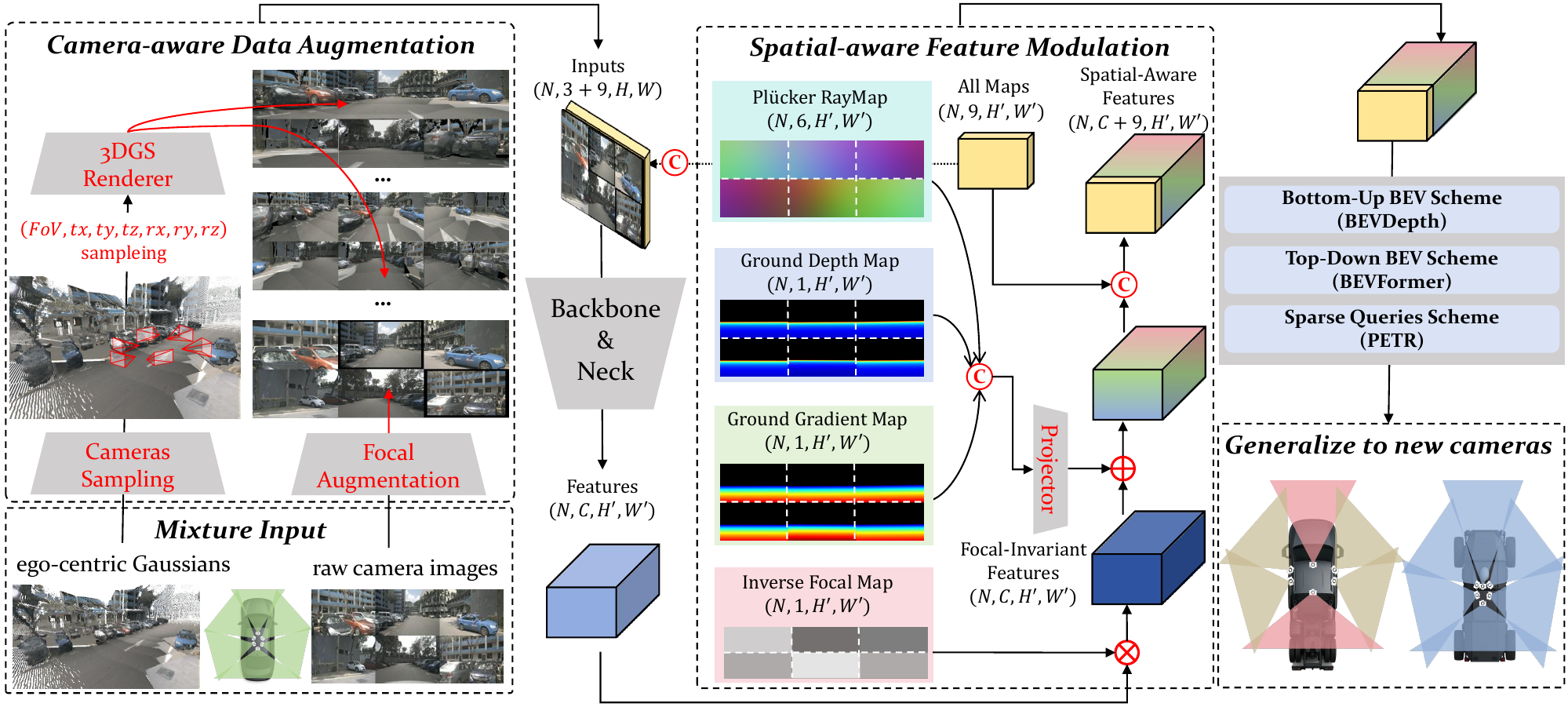}
  \caption{
  Illustration of our CoIn3D framework for generalizable MC3D across multi-camera configurations. During training, we apply the camera-aware data augmentation (CDA) to generate $N$ images with randomly sampled camera configurations, followed by spatial-aware feature modulation (SFM). SFM modulates activations using an inverse focal map to obtain focal-invariant features, then projects prior maps (ground depth, gradient map, and Plücker raymap) to create spatial embeddings, which are added to the focal-invariant features. These maps are concatenated with image input and features to provide raw priors. Finally, the spatial-aware features can be easily integrated into MC3D for downstream tasks. During inference, we use raw images and apply spatial-aware modulation to generalize to new camera configurations. Our framework is applicable to dominant MC3D paradigms, including bottom-up BEV, top-down BEV, and sparse-queries.
  } 
  \label{fig:main-framework}
  \vspace{-10pt}
\end{figure*}

\subsection{Multi-Camera 3D Object Detection}
Current MC3D models fall into three paradigms.

\textbf{Bottom-up BEV schemes} \cite{roddick2018orthographic,philion2020lift,huang2021bevdet,huang2022bevdet4d,li2023bevdepth,yang2025bevheight++} splat the image features into BEV feature, and then objects are detected on the BEV feature.
Among them, BEVDepth \cite{li2023bevdepth} and BEVDet4D \cite{huang2022bevdet4d} are two representative schemes, which consider LiDAR depth supervision and temporal fusion.

\textbf{Top-down BEV schemes}, represented by BEVFormer \cite{li2024bevformer,yang2023bevformer}, update BEV query features by applying spatial cross-attention with image features and temporal self-attention with adjacent BEV features. Then, the detection results will be decoded on BEV features.

\textbf{Sparse queries schemes} \cite{wang2022detr3d,liu2022petr,wang2023exploring,park2022time,liu2023sparsebev} encode 3D scenes into sparse representations. Among them, PETR \cite{liu2022petr} injects 3D position embeddings into image features and StreamPETR \cite{wang2023exploring} explores long sequence modeling.

These MC3D paradigms have their own advantages. While BEV feature-based schemes can provide dense scene representation for holistic 3D understanding, sparse queries offer lower computational costs. In this paper, we seek a unified framework to improve the generalization of all these dominant paradigms across camera configurations.

\subsection{Camera Configuration Generalization}

To tackle the generalization problem caused by the configuration gap, on one hand, some works try to adjust the image input \cite{yang2024geometry,zheng2023cross,li2024unidrive}, and on the other hand, some works try to adjust the predicted target \cite{wang2023towards,lu2025towards,chang2024unified,kumar2025charm3r}. 

\textbf{For image input adjustment}, \cite{yang2024geometry,zheng2023cross,li2024unidrive} resize and crop the image to align focal length and FoV. To adjust the extrinsic gap, \cite{yang2024geometry} and \cite{li2024unidrive} warp image based on spherical and cylindrical space assumption, respectively. However, resizing image will cause image shape misalignment and cannot take advantage of batch acceleration. Resizing long-focal images to short focal lengths will result in a loss of resolution. Cropping image will loss texture information. Warping images based on spherical or cylindrical space assumption will destroy the 3D scene structure.

\textbf{For prediction adjustment}, DG-BEV, PD-BEV and UDGA-BEV \cite{wang2023towards,lu2025towards,chang2024unified} treat all cameras as having an identical virtual focal length in both the training and testing data, and rescale the virtual depth to real depth according to the ratio between virtual and real focal lengths. PD-BEV further constrains feature consistency between image features and BEV feature, while UDGA-BEV further constrains depth and photometric consistency between different cameras to enhance the generalization of models. \cite{kumar2025charm3r} introduces the ground depth of the object bottom center to help model overcome camera height gap. However, these methods do not fully and explicitly consider camera configurations. Furthermore, these methods cannot apply to all MC3D paradigms due to their depth-based design.

In this paper, we comprehensively revisit the influence of multi-camera configurations on MC3D and then propose CoIn3D to uniformly mitigate the camera configuration gap. Our scheme can be applied to all MC3D paradigms.

\vspace{-5pt}
\subsection{3D Gaussian splatting}
Novel view synthesis \cite{klinghoffer2023towards} can augment images under different configurations, showing the potential to achieve configuration-invariant MC3D. 3DGS \cite{kerbl20233d} has a fast rendering speed and makes dynamic augmentation possible. However, current 3DGS schemes \cite{yan2024street,chen2024omnire,wei2025omni,jiang2025anysplat,lu2024drivingrecon} have high training costs. Unlike previous works, we propose a cost-efficient, training-free ego-centric Gaussian construction pipeline tailored for MC3D data augmentation.

%% file: sec/3_revisit.tex
\begin{figure}[t]
  \centering
   \includegraphics[width=1.0\linewidth]{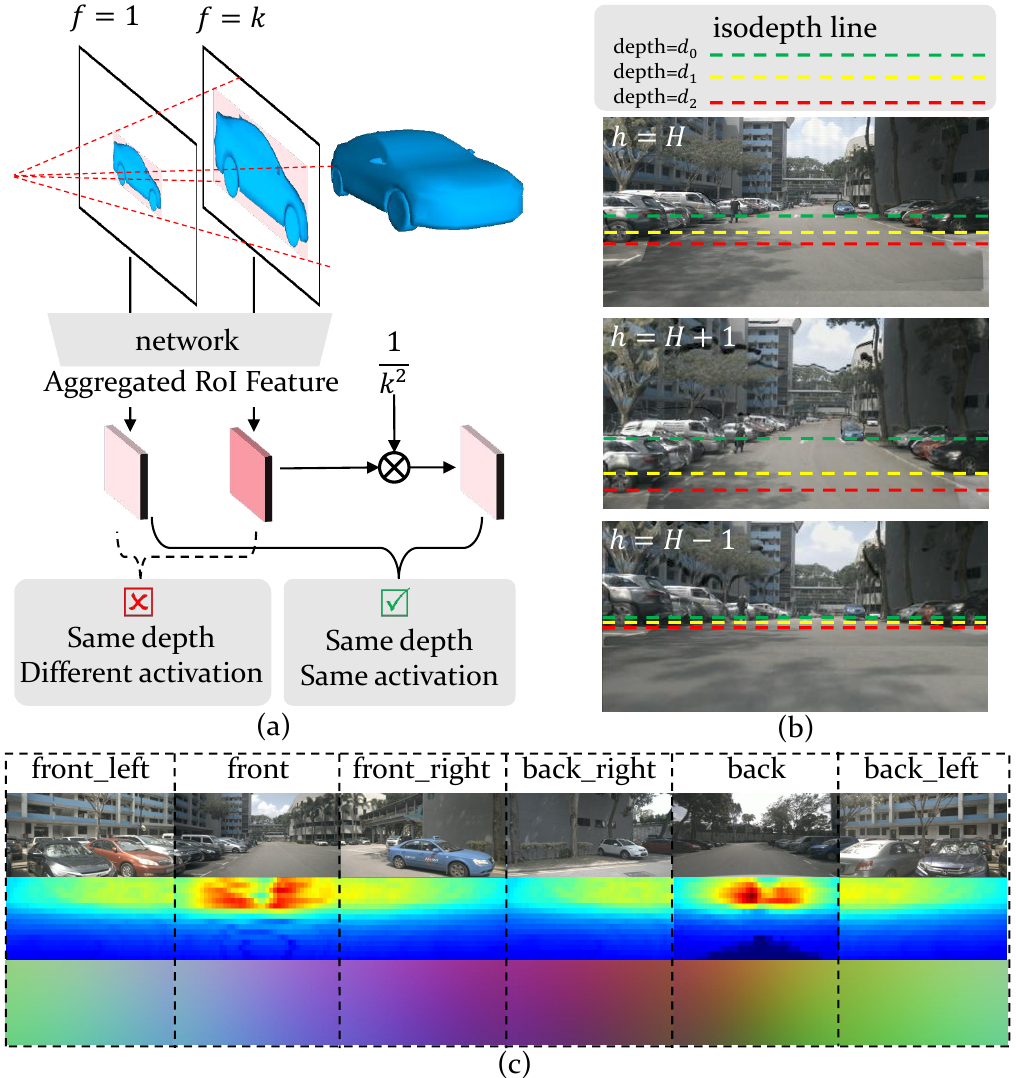}
   \caption{Illustration of the spatial discrepancies under different camera configurations: (a) focal-ambiguity for a same object; (b) ground depth and depth increasing rate under different camera heights; (c) the scene structure (1st row), depth distribution (2nd row), and Plücker raymap (3rd row) for surround-view cameras.}
   \label{fig:prior-explanaion}
   \vspace{-5pt}
\end{figure}

\section{Revisit cameras configuration in MC3D}
\label{sec:revisit}

\subsection{MC3D task}
MC3D takes $N$ surrounding camera images $I=\{i_1,i_2,...,i_N\} \in \mathbb{R}^{N\times 3\times H\times W}$ as input. Each camera has an intrinsic matrix $\mathbf{K}\in \mathbb{R}^{3\times3}$ and an extrinsic matrix $\mathbf{T}=[\mathbf{R};\mathbf{t}|\mathbf{0},1] \in \mathbb{R}^{4 \times4}$, which can transform points from the camera system to the ego system. Taking all these as input, MC3D aims to localize and categorize objects in 3D ego space. MC3D needs to predict position $(x,y,z)$, size $(l,w,h)$, orientation $\theta$, and category of each object.

\subsection{Intrinsic revisiting}
\label{sec:revisit_intrinsic}

Different cameras has different focal length and field-of-view (FoV). For focal length, as shown in Fig. \ref{fig:prior-explanaion}(a), the same object under different focal plane has different sizes, which hinders the model's consistent depth understanding.
For FoV, different FoVs corresponds to different scene perspective geometries. For example, given a camera height $H_{cam}$, the initial ground depth $z_i$ on image bottom is determined by the FoV angle $\alpha$: $z_i=\frac{H_{cam}}{\tan(\frac{\alpha}{2})}$. This prior can help model understand scene structure.

\begin{figure}[t]
  \centering
  \includegraphics[width=1.0\linewidth]{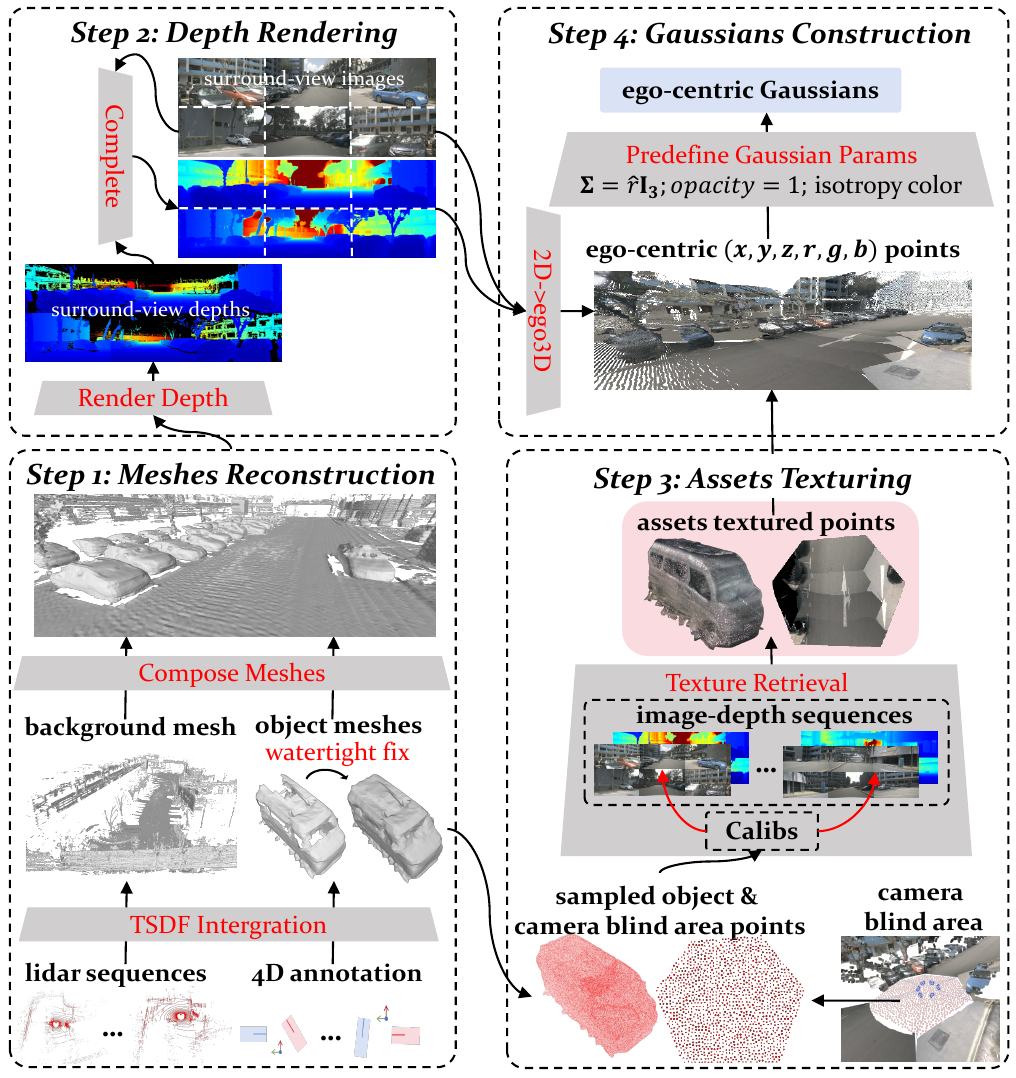}
  \caption{Illustration of the training-free ego-centric Gaussians construction pipeline. We transform reconstructed texture point clouds into Gaussian representations using predefined parameters.}
  \label{fig:training-free-3dgs}
  \vspace{-5pt}
\end{figure}

\subsection{Extrinsic revisiting}
Different camera mounting positions and orientations results in varying extrinsics and spatial priors.

First, ground plane can provide a strong prior for MC3D. We assume the ground plane is flat, so we can use $N$ non-collinear points ($N \geq 3$) $\textbf{p}_e \in \mathbb{R}^{N \times 4}$ on the ground to define a plane in ego space. The ground plane equation $Ax+By+Cz+D=0$ in each camera space can be solved by Least Squares from points $\textbf{p}_c=\textbf{T}^{-1}\textbf{p}_e$. For an image with intrinsic parameters $f_u, f_v, c_u, c_u$, if we assume that a pixel $(u, v)$ with depth $z(u,v)$ is on the ground, then the depth can be solved by:
\begin{equation}
    z(u,v) = -\frac{D}{AX+BY+C}
    \label{eq:ground_depth}
\end{equation}
where $X=\frac{u-c_u}{fu}$ and $Y=\frac{v-c_v}{f_v}$.

Different camera mounting heights also lead to varying perspective effects. As shown in Fig.\ref{fig:prior-explanaion}(b), if we observe the ground in the image from near to far, we will find that, higher camera mounting height has slower ground depth increase rate, while lower height causes a faster increase. Training on one specific camera height will cause overfitting to this rate change behavior.

Finally, different camera orientations result in different observed scene geometry. As shown in Fig. \ref{fig:prior-explanaion}(c), each camera captures different depth distribution and scene structures, which should be considered as priors.

\subsection{Array revisiting}
Different platforms will adopt camera arrays with different numbers of cameras and cameras layouts, resulting in different surround-view structures. As shown in Fig. \ref{fig:prior-explanaion}(c), each surround-view image can be seen as a fragment of the omnidirectional image of the scene. Therefore, the scene structure is continuous across different images. Moreover, different layouts result in different overlap regions. These two factors should be considered by MC3D during multi-camera correlation and feature fusion.

%% file: sec/4_methodology.tex
\section{Methodology}
\label{sec:methodology}

We identify that discrepancies in intrinsics, extrinsics, and array layouts are the key challenge for cross-configuration generalization. To address this, we propose CoIn3D, a framework for generalizing MC3D across configurations.

The framework is shown in Fig. \ref{fig:main-framework}. During training, taking raw images and ego-centric Gaussians as input, we first apply camera-aware data augmentation (CDA) to dynamically generate images with diverse camera configurations. After applying CDA, the augmented images $\mathbf{I}_{im} \in \mathbb{R}^{3 \times H \times W}$ are fed into the backbone and neck to extract features $\mathbf{F}_{im} \in \mathbb{R}^{C \times H' \times W'}$. We then apply spatial-aware feature modulation (SFM) to enrich the image feature by embedding four spatial prior maps formulating configurations. Finally, the modulated spatial-aware feature can be integrated into any MC3D paradigm for downstream tasks. During inference, the model only takes raw images as input and applies SFM to generalize to new camera configurations.

\subsection{Spatial-aware feature modulation}
\label{sec:method_GFM}
In this section, we introduce how we modulate the raw image features to obtain spatial-aware features.

\subsubsection{Inverse focal map}
\label{sec:method_GFM_InvF}

We first address focal ambiguity by assuming that features with different focal lengths should have similar activations. Larger focal lengths can be seen as upsampled versions of smaller ones, leading to different aggregated RoI feature activations for the same object, which causes ambiguity.

To eliminate the ambiguity, we propose using the square of the focal length to normalize the feature activation, which is based on the observation that a $k$-fold difference in focal length leads to a $k^2$-fold difference in pixel size for the same object. Specifically, for a raw image feature $\textbf{F}_{im}$ with focal length $f$, we use the inverse of the square focal map $\textbf{M}_{IF}\in \mathbb{R}^{1 \times H'\times W'}$ to normalize as follows:
\begin{equation}
    \begin{aligned}
        \textbf{M}_{IF}=\mathbf{1}*\frac{1}{f^2}\\
        \textbf{F}_{im}^{1} = \textbf{M}_{IF} \odot \textbf{F}_{im}
    \end{aligned}
    \label{eq:M_invF}
\end{equation}

\subsubsection{Ground depth and gradient map}
\label{sec:method_GFM_GDGG}
After addressing the focal gap, we consider the ground prior under different camera configurations. The ground depth map $\textbf{M}_{GD} \in \mathbb{R}^{1 \times H'\times W'}$ can be generated by Eq. \ref{eq:ground_depth} and provides a straightforward scene spatial prior. 

We further propose the ground gradient map. It is based on the observation that, under different camera mounting heights, the ground depth increase rating from near to far will be quite different as shown in Fig. \ref{fig:prior-explanaion} and results in different perspective effects. To formulate, we introduce ground gradient map $\textbf{M}_{GG} \in \mathbb{R}^{1 \times H\times W}$, which can be derived from $\textbf{M}_{GD}$ through cross-row difference: $\textbf{M}_{GG}=\textbf{M}_{GD}[:, :-1,:]-\textbf{M}_{GD}[:, 1:, :]$. However, the numerical values of the results are small and vary rapidly, which is unfavorable for network training. To make the numerical value suitable for network, we apply a log-inverse transformation to get the final ground gradient map:
\begin{equation}
    \textbf{M}_{GG} = log(\frac{1}{\textbf{M}_{GD}[:, :-1,:]-\textbf{M}_{GD}[:, 1:, :]} +1)
    \label{eq:M_GG}
\end{equation}

\subsubsection{Plücker raymap}
\label{sec:method_GFM_PR}

To further enriching the spatial representation, we introduce the Plücker raymap inspired by \cite{zhang2024cameras,plucker1828analytisch}. The Plücker raymap describes the direction and moment of a cluster of rays emitted from the optical center of camera to each pixel.

Specifically, for a camera with an intrinsic matrix $\mathbf{K} \in \mathbb{R}^{3 \times 3}$ and extrinsic matrix, which describes the transformation from camera to ego $\mathbf{T}= [\textbf{R},\textbf{t};\textbf{0},1] \in \mathbb{R}^{4 \times 4}$, we can construct the Plücker raymap as follows. 
For a pixel, denoted its homogeneous coordinates as $\mathbf{p}=[u,v,1]^T$, its ray direction $\mathbf{d} \in \mathbb{R}^3$ in ego and ray moment related to the camera origin $\mathbf{m} \in \mathbb{R}^3$ can be computed as follows:
\begin{equation}
    \begin{aligned}
        \mathbf{d}=\mathbf{R}\mathbf{K}^{-1}\mathbf{p}\\
        \mathbf{m}=\mathbf{t} \times \mathbf{d}
    \end{aligned}
    \label{eq:M_PR}
\end{equation}
where $\mathbf{d}$ together with $\mathbf{m}$ construct the Plücker coordinate $\mathbf{r}=(\mathbf{d}, \mathbf{m}) \in \mathbb{R}^6$ of each pixel in the Plücker raymap $\textbf{M}_{PR} \in \mathbb{R}^{6 \times H' \times W'}$. 

The Plücker raymap has several beneficial properties. First, it formulates camera's FoV, rotation, and translation in the ego system. Second, it continuously formulates the pixel position among multi-camera images which can utilized for camera correlation and feature fusion.

\subsubsection{Spatial embedding module}
The ground depth map, gradient map and Plücker raymap together provide an over-complete representation that describes camera configurations. To embed these prior maps into the feature, similar to positional embeddings in transformers \cite{vaswani2017attention}, we first concatenate all these prior maps channel-wise. Then we use a shallow projector to project it to higher-level feature space to obtain spatial embedding. Finally, we add spatial embedding to the image feature. This process can be described as follows:
\begin{equation}
    \textbf{F}_{im}^2 = \textbf{F}_{im}^1 + projector(cat(\textbf{M}_{GD}, \textbf{M}_{GG}, \textbf{M}_{PR}))
    \label{eq:M_MixPriorProj}
\end{equation}

In our implementation, the projector has a simple architecture with a 3 $\times$ 3 convolution and a ReLU activation layer. Finally, all prior maps are concatenated with the modulated feature $\textbf{F}_{im}^3=cat(\textbf{M}_{IF},\textbf{M}_{GD},\textbf{M}_{GG},\textbf{M}_{PR},\textbf{F}_{im}^2)$ and image input to provide raw prior information. The spatial-aware feature $\textbf{F}_{im}^3$ can be used by all kinds of MC3D schemes to decode the 3D detection results.

\subsection{Camera-aware data augmentation}
\label{sec:method_GDA}
Data augmentation is important for enhancing the generalization of MC3D. However, basic augmentations like flipping and photometric distortion cannot augment the camera configuration. To address it, we propose a training-free 3DGS scheme to construct the ego-centric Gaussians for dynamic data augmentation in a cost-efficient manner.

\subsubsection{Training-free 3DGS construction}
\label{sec:method_GDA_3dgs}
Specifically, the training-free 3DGS construction pipeline can be divided into four steps, as shown in Fig. \ref{fig:training-free-3dgs}. First, we use 4D annotations to decompose LiDAR sequences into background and objects. Next, we use TSDF integration \cite{vizzo2022vdbfusion} to reconstruct the background mesh and object meshes. Moreover, we fix the object meshes to watertight surface.

Second, for each annotated timestamp in the sequence, we compose the background mesh and objects meshes according to calibration and annotations. Next, we render the composed meshes to depth \cite{ravi2020accelerating} for each camera. These mesh-rendered depths have an advantage that their metric is naturally precise and they can ensure surround-view consistency. These merits are well-suited for MC3D data augmentation. Finally, we apply depth completion \cite{wang2024scale} to fill in holes without meshes and get dense depths.

Third, we reconstruct auxiliary assets, including the texture point model of objects and camera blind area. These assets help complete the unseen parts beyond raw imaging. Specifically, we first sample points from object meshes and camera blind area. Then we warp and retrieve textures through depth matching across images in the sequence.

Fourth, we construct the ego-centric Gaussians for each annotated data frame. We first project RGB-D images to ego space to obtain texture point cloud. Then, we append the texture assets. Finally, we set the covariance matrix of Gaussian as $\mathbf{\Sigma}=\hat{r}\mathbf{I}_3$, which means that each Gaussian has no rotation and has a fixed predefined radius. The opacity of each Gaussian is set to 1. We set an isotropy color for each Gaussian, which meaning that we use the raw $(r,g,b)$ color, the center of Gaussian is determined by the $(x,y,z)$ coordinates of the point clouds. In other words, our scheme can be seen as a point-rendering scheme but we transform point cloud into Gaussian representation to take advantage of the fast rendering speed of 3DGS. The ego-centric Gaussians can be rendered at about 450 fps. We provide more details about this pipeline in the supplementary material.

\subsubsection{Data augmentation}
\label{sec:method_GDA_aug}
We will apply our data augmentation to both the ego-centric Gaussians and raw image inputs. These data will be dynamically augmented during training.

Specifically, for ego-centric Gaussians, we randomly sample new camera configurations to render novel-view images. For raw images, we randomly resize them according to the principal point and adjust the focal length. 

%% file: sec/5_experiments.tex
\section{Experiment}
\label{sec:experiment}

\input{tables/main_exp_bevdepth_compact}

\input{tables/main_exp_bevformer_petr_compact}

We conduct comprehensive experiments across multiple landmark datasets based on different MC3D paradigms. 

\subsection{Experiment Setup}
\textbf{Datasets.} We conduct experiments across three landmark datasets, including Nuscenes \cite{caesar2020nuscenes}, Waymo \cite{sun2020scalability}, and Lyft \cite{3d-object-detection-for-autonomous-vehicles}. These datasets are collected with quite different camera configurations, making them challenging for evaluating the generalization ability of MC3D across configurations. Our model is trained on one dataset and tested on the others for each experiment. Detailed dataset information is provided in the supplementary material.

\noindent\textbf{Evaluation Metrics.} Following previous work \cite{wang2023towards,lu2025towards,chang2024unified}, we adopt the metric NDS$^*$ that aggregates mean Average Precision (mAP), mean Average Translation Error (mATE), mean Average Scale Error (mASE), and mean Average Orientation Error (mAOE):
\begin{equation}
    \text{NDS}^*=\frac{1}{6}[3\text{mAP}+\sum_{\text{mTP}\in \mathbb{TP}}(1-\text{min}(1, \text{mTP}))]
    \label{eq:nds*}
\end{equation}

Following previous work \cite{chang2024unified}, we consider the unified category ``car'' for generalization experiment. Moreover, the ``car", ``truck", ``construction vehicle", ``bus" and ``trailer" in Nuscenes will all be considered as ``car", except for the Lyft $\rightarrow$ Nuscenes evaluation. We only validate results in the range $[-50m, 50m]$ during training and validation.

\noindent\textbf{Implementation Details.} We adopt BEVDepth-R50 \cite{li2023bevdepth}, BEVFormer-tiny \cite{li2024bevformer} and PETR-vov \cite{liu2022petr} as base models to represent the three MC3D paradigms, respectively. We train the model on 4 RTX 4090 GPUs. The image width will be downsampled to 704 for BEVDepth and 800 for BEVFormer and PETR. The image height will be downsampled by the same ratio as the width. More implementation details are shown in the supplementary material.

\subsection{Main Results}
\subsubsection{Results based on BEVDepth}

Tab. \ref{table:main_exp_bevdepth_compact} verifies the effectiveness of our scheme on bottom-up BEV framework (e.g., BEVDepth-R50). Our scheme significantly enhances the generalization of BEVDepth across datasets with different camera configurations compared to the base model. Moreover, our scheme achieves SOTA performance compared to current methods.

Specifically, directly transfer BEVDepth to new configuration leads to poor performance. Especially in Nuscenes$\rightarrow$Waymo and Waymo$\rightarrow$Nuscenes, the mAP drops to nearly 0 because these two datasets have a huge camera configuration gap. With our framework, the generalization performance NDS$^*$ gap is bridged from 0.178 to 0.513, 0.296 to 0.534, 0.133 to 0.481, and 0.213 to 0.452 for Nuscenes$\rightarrow$Waymo, Nuscenes$\rightarrow$Lyft, Waymo$\rightarrow$Nuscenes and Lyft$\rightarrow$Nuscenes settings, respectively.

Further, we compare the performance with current methods: CAM-Convs \cite{facil2019cam}, Single-DGOD \cite{wu2022single}, DG-BEV \cite{wang2023towards}, PD-BEV \cite{lu2025towards} and UDGA-BEV \cite{chang2024unified}. CAM-Convs and Single-DGOD are applied in depth prediction and 2D object detection. DG-BEV, PD-BEV and UDGA-BEV are three current SOTA schemes. We observe that, CAM-Convs and Single-DGOD have little effect because they did not consider the camera configuration problem in MC3D. DG-BEV, PD-BEV, and UDGA-BEV make great progress by considering the focal ambiguity, feature consistency and depth consistency. 

In this paper, by fully and explicitly considering camera configuration, we surpass all previous methods and achieve SOTA performance. Specifically, we surpass UDGA-BEV, the SOTA scheme, by 0.054, 0.047, 0.004 and 0.031NDS$^*$ under Nuscenes$\rightarrow$Waymo, Nuscenes$\rightarrow$Lyft, Waymo$\rightarrow$Nuscenes and Lyft$\rightarrow$Nuscenes settings.

\input{tables/main_ablation}

\vspace{-5pt}
\subsubsection{Results based on BEVFormer and PETR}

Tab. \ref{table:main_exp_bevformer_petr(n2w)_compact} verifies that our model-agnostic design enables application on both top-down BEV (e.g., BEVFormer-tiny) and sparse query-based frameworks (e.g., PETR-vov), which were not supported by most previous methods. Notably, PD-BEV did not mention the versions of BEVFormer and PETR used, so we choose the smallest versions for these two base models. Overall, our framework is a unified solution for all MC3D paradigms.

\subsection{Ablation Studies}
In this section, we study the effect of the two modules, SFM and CDA, in our scheme through BEVDepth. All experiments are conducted on Nuscenes$\rightarrow$Waymo setting because we consider that they have the largest configuration gap.

\subsubsection{Main Ablations}

Tab. \ref{table:main_ablation} demonstrates the effectiveness of the proposed SFM and CDA. CA represents the camera-aware SE-module designed by BEVDepth, which can embed camera configuration in a simple way. The results show that our SFM can work without CDA, as it explicitly formulates the camera configuration. Applying CDA alone is ineffective due to the discrepancies of configurations that has not been processed. Combining SFM and CDA can promote NDS$^*$ from 0.178 to 0.513 and surpass CA and CDA by 0.166. This means that, on one hand, our framework can greatly improve the generalization performance across configurations, on the other hand, SFM provides a more effective way to formulate configuration. In addition, combining CA with our scheme leads to degraded performance because CA disturbs the features we modulate. Therefore, we discard CA.

\subsubsection{Ablation on SFM}

Tab. \ref{table:GFM_ablation} demonstrates the effectiveness of all four spatial priors in our scheme based on CDA, including the inverse focal map (IF), ground depth map (GD), ground gradient map (GG), and Plücker raymap (PR).

First, using IF alone improves NDS$^*$ by 0.238, showing that inverse focal modulation effectively mitigates focal ambiguity by normalizing feature activation. Second, adding GD brings a 0.036 gain, as the explicit ground depth prior helps the model leverage the crucial ground-plane prior. Third,  adding GG also brings an additional 0.008 gain, indicating that over-complete ground-related priors enhance spatial features. Finally, adding PR further brings a 0.007 gain, demonstrating its ability to encode pixel-level spatial priors for camera configuration.

\input{tables/GFM_ablation}

\input{tables/GDA_ablation_v2}

\subsubsection{Ablation on CDA}

To analyze the effect of each augmentation, we apply focal augmentation (F-Aug) and our novel view synthesis augmentation (NVS-Aug) based on SFM in Tab. \ref{table:GDA_ablation_v2}. F-Aug improves NDS$^*$  by 0.060, indicating that SFM benefits from augmentation. NVS-Aug boosts NDS* by 0.095, showing that the effect of focal augmentation on raw images is limited, while NVS-Aug better enhances generalization by augmenting training data with diverse configurations.

%% file: tables/main_exp_bevdepth_compact.tex
\begin{table*}[t]
\caption{Comparison of our scheme with existing methods based on BEVDepth. ``Nuscenes $\rightarrow$ Waymo'' refers to train on Nuscenes and test on Waymo with different configurations, and similarly for the other datasets.  ``Direct Transfer'' refers to directly use BEVDepth to test on target dataset without any adaptation. ``Oracle'' refers to training and testing on target dataset, serving as an upper bound for performance.}
\centering

\begin{tabular}{@{}c|ccccc|ccccc@{}}
\toprule
Setting         & \multicolumn{5}{c|}{Nuscenes $\rightarrow$ Waymo}                                                & \multicolumn{5}{c}{Waymo $\rightarrow$ Nuscenes}                                                 \\ \midrule
Method          & mAP$\uparrow$            & mATE$\downarrow$           & mASE$\downarrow$           & mAOE$\downarrow$           & \textbf{NDS*}$\uparrow$           & mAP$\uparrow$            & mATE$\downarrow$           & mASE$\downarrow$           & mAOE$\downarrow$           & \textbf{NDS*}$\uparrow$           \\ \midrule
Oracle          & 0.552          & 0.528          & 0.148          & 0.085          & 0.649          & 0.475          & 0.577          & 0.177          & 0.147          & 0.587          \\
Direct Transfer & 0.040          & 1.303          & 0.265          & 0.790          & 0.178          & 0.032          & 1.305          & 0.768          & 0.532          & 0.133          \\
CAM-Convs\cite{facil2019cam}       & 0.045          & 1.301          & 0.253          & 0.773          & 0.185          & 0.038          & 1.308          & 0.316          & 0.506          & 0.215          \\
Single-DGOD\cite{wu2022single}     & 0.034          & 1.305          & 0.262          & 0.855          & 0.164          & 0.014          & 1.000          & 1.000          & 1.000          & 0.007          \\
DG-BEV\cite{wang2023towards}          & 0.297          & 0.822          & \textbf{0.216} & 0.372          & 0.415          & 0.303          & 0.689          & \textbf{0.218} & 0.171          & 0.472          \\
UDGA-BEV\cite{chang2024unified}        & 0.349          & 0.754          & 0.289          & 0.250          & 0.459          & 0.326          & \textbf{0.684} & 0.263          & \textbf{0.168} & 0.477          \\ \midrule
\textbf{Ours}   & \textbf{0.381} & \textbf{0.687} & 0.220          & \textbf{0.155} & \textbf{0.513} & \textbf{0.349} & 0.727          & 0.257          & 0.179          & \textbf{0.481} \\ \midrule
Setting         & \multicolumn{5}{c|}{Nuscenes $\rightarrow$ Lyft}                                                 & \multicolumn{5}{c}{Lyft $\rightarrow$ Nuscenes}                                                  \\ \midrule
Method          & mAP$\uparrow$            & mATE$\downarrow$           & mASE$\downarrow$           & mAOE$\downarrow$           & \textbf{NDS*}$\uparrow$           & mAP$\uparrow$            & mATE$\downarrow$           & mASE$\downarrow$           & mAOE$\downarrow$           & \textbf{NDS*}$\uparrow$           \\ \midrule
Oracle          & 0.602          & 0.471          & 0.152          & 0.078          & 0.684          & 0.475          & 0.577          & 0.177          & 0.147          & 0.587          \\
Direct Transfer & 0.112          & 0.997          & 0.176          & 0.389          & 0.296          & 0.102          & 1.143          & 0.239          & 0.789          & 0.213          \\
CAM-Convs       & 0.145          & 0.999          & 0.173          & 0.368          & 0.316          & 0.098          & 1.198          & 0.209          & 1.064          & 0.181          \\
Single-DGOD     & 0.159          & 0.949          & 0.174          & 0.358          & 0.332          & 0.105          & 1.166          & 0.222          & 0.905          & 0.198          \\
DG-BEV          & 0.287          & 0.771          & 0.170          & 0.302          & 0.437          & 0.268          & 0.764          & 0.205          & 0.591          & 0.374          \\
PD-BEV\cite{lu2025towards}          & 0.304          & 0.709          & 0.169          & 0.289          & 0.458          & 0.263          & 0.746          & 0.186          & 0.790          & 0.344          \\
UDGA-BEV        & 0.324          & 0.709          & 0.162          & 0.180          & 0.487          & 0.281          & 0.759          & 0.183          & 0.377          & 0.421          \\ \midrule
\textbf{Ours}   & \textbf{0.375} & \textbf{0.660} & \textbf{0.161} & \textbf{0.101} & \textbf{0.534} & \textbf{0.303} & \textbf{0.647} & \textbf{0.176} & \textbf{0.377} & \textbf{0.452} \\ \bottomrule
\end{tabular}

\label{table:main_exp_bevdepth_compact}
\end{table*}

%% file: tables/main_exp_bevformer_petr_compact.tex
\begin{table*}[t]
\caption{Generalization experiment of our scheme based on BEVFormer and PETR. Our framework can apply to both top-down BEV and sparse queries schemes, represented by BEVFormer and PETR respectively. Notably, most previous methods do not support these two paradigms. N, L, W, and DT represent Nuscenes, Lyft, Waymo, and Direct Transfer, respectively.}
\centering

\begin{tabular}{@{}c|c|ccccc|ccccc@{}}
\toprule
\multirow{2}{*}{Setting} & \multirow{2}{*}{Method} & \multicolumn{5}{c|}{BEVFormer}                                                     & \multicolumn{5}{c}{PETR}                                                           \\ \cmidrule(l){3-12} 
                         &                         & mAP$\uparrow$            & mATE$\downarrow$            & mASE$\downarrow$           & mAOE$\downarrow$           & \textbf{NDS*}$\uparrow$            & mAP$\uparrow$             & mATE$\downarrow$           & mASE$\downarrow$           & mAOE$\downarrow$           & \textbf{NDS*}$\uparrow$            \\ \midrule
\multirow{3}{*}{N $\rightarrow$ L}     & DT                      & 0.149          & 1.031          & 0.755          & 1.241          & 0.115          & 0.013          & 1.240          & 0.761          & 1.484          & 0.046          \\
                         & PD-BEV                  & 0.208          & /              & /              & /              & 0.355          & 0.032          & /              & /              & /              & 0.091          \\
                         & \textbf{Ours}           & \textbf{0.237} & \textbf{0.850} & \textbf{0.191} & \textbf{0.407} & \textbf{0.377} & \textbf{0.332} & \textbf{0.880} & \textbf{0.170} & \textbf{0.210} & \textbf{0.456} \\ \midrule
\multirow{2}{*}{N $\rightarrow$ W}     & DT                      & 0.054          & 1.146          & 0.747          & 1.606          & 0.069          & 0.048          & 1.191          & 0.748          & 1.543          & 0.066          \\
                         & \textbf{Ours}           & \textbf{0.249} & \textbf{0.872} & \textbf{0.239} & \textbf{0.365} & \textbf{0.379} & \textbf{0.200} & \textbf{1.141} & \textbf{0.242} & \textbf{0.505} & \textbf{0.309} \\ \bottomrule
\end{tabular}

\label{table:main_exp_bevformer_petr(n2w)_compact}
\vspace{-5pt}
\end{table*}

%% file: tables/main_ablation.tex
\begin{table}[]
\caption{Main ablation of our two modules CDA and SFM. CA refers the Camera-Aware SE module designed by BEVDepth.}
\centering

\begin{tabular}{@{}ccc|cccc@{}}
\toprule
\multicolumn{3}{c|}{Module} & \multirow{2}{*}{mAP$\uparrow$} & \multirow{2}{*}{mATE$\downarrow$} & \multirow{2}{*}{mAOE$\downarrow$} & \multirow{2}{*}{\textbf{NDS*}$\uparrow$} \\
CDA      & SFM     & CA     &                      &                       &                       &                       \\ \midrule
         &         & \checkmark      & 0.040                & 1.303                 & 0.790                 & 0.178                 \\
\checkmark        &         & \checkmark      & 0.190                & 1.021                 & 0.255                 & 0.347                 \\
\checkmark        & \checkmark       & \checkmark      & 0.381                & 0.706                 & 0.168                 & 0.504                 \\ \midrule
         & \checkmark       &        & 0.215                & 0.887                 & 0.312                 & 0.358                 \\
\checkmark        &         &        & 0.041                & 1.310                 & 0.513                 & 0.224                 \\
\checkmark        & \checkmark       &        & \textbf{0.381}       & \textbf{0.687}        & \textbf{0.155}        & \textbf{0.513}        \\ \bottomrule
\end{tabular}

\label{table:main_ablation}
\vspace{-10pt}
\end{table}

%% file: tables/GFM_ablation.tex
\begin{table}[]
\caption{Ablation on our SFM based on CDA. IF, GD, GG, and PR represent the inverse focal map, ground depth map, ground gradient map, and Plücker raymap, respectively.}
\centering

\begin{tabular}{@{}cccc|cccc@{}}
\toprule
\multicolumn{4}{c|}{Spatial Priors} & \multirow{2}{*}{mAP$\uparrow$} & \multirow{2}{*}{mATE$\downarrow$} & \multirow{2}{*}{mAOE$\downarrow$} & \multirow{2}{*}{\textbf{NDS*}$\uparrow$} \\
IF  & GD  & GG         & PR         &                      &                       &                       &                       \\ \midrule
    &     &            &            & 0.041                & 1.310                 & 0.513                 & 0.224                 \\
\checkmark   &     &            &            & 0.332                & 0.811                 & 0.176                 & 0.462                 \\
\checkmark   & \checkmark   & \textbf{}  & \textbf{}  & 0.365                & 0.696                 & 0.182                 & 0.498                 \\
\checkmark   & \checkmark   & \checkmark          &            & 0.375                & \textbf{0.684}        & 0.177                 & 0.506                 \\
\checkmark   & \checkmark   & \checkmark          & \checkmark          & \textbf{0.381}       & 0.687                 & \textbf{0.155}        & \textbf{0.513}        \\ \bottomrule
\end{tabular}

\label{table:GFM_ablation}
\end{table}

%% file: tables/GDA_ablation_v2.tex
\begin{table}[]
\caption{Ablation on our CDA based on SFM. F-Aug and NVS-Aug represent focal augmentation and novel view synthesis augmentation, respectively.}
\centering

\begin{tabular}{@{}cc|cccc@{}}
\toprule
\multicolumn{2}{c|}{Augmentation} & \multirow{2}{*}{mAP$\uparrow$} & \multirow{2}{*}{mATE$\downarrow$} & \multirow{2}{*}{mAOE$\downarrow$} & \multirow{2}{*}{\textbf{NDS*}$\uparrow$} \\
F-Aug          & NVS-Aug          &                      &                       &                       &                       \\ \midrule
               &                  & 0.215                & 0.887                 & 0.312                 & 0.358                 \\
\checkmark              &                  & 0.314                & 0.990                 & 0.192                 & 0.418                 \\
\checkmark              & \checkmark                & \textbf{0.381}       & \textbf{0.687}        & \textbf{0.155}        & \textbf{0.513}        \\ \bottomrule
\end{tabular}

\label{table:GDA_ablation_v2}
\end{table}

%% file: sec/6_conclusion.tex
\section{Conclusion}
\label{sec:conclusion}
In this paper, we revisit how different camera configurations affect the generalization of multi-camera 3D object detection, and we identify that the devil lies in spatial prior discrepancies across source and target configurations. To address this, we propose CoIn3D, a generalizable MC3D framework that enables strong transferability from source configurations to unseen target ones. Extensive experiments across three landmark datasets with different camera configurations show that our framework is effective for dominant MC3D paradigms. We hope that this work can bring insight for industrial applications to facilitate deployment. In the future, we will explore improving the generalization performance of MC3D across different semantic distributions.

%% file: sec/7_ack.tex
\section*{Acknowledgments}
This work was supported by the New Generation Artificial Intelligence-National Science and Technology Major Project under Grant 2025ZD0124203,
NSFC under Grants 62373298 and U24A20252, Guangdong Pearl River Talent Program under Grant 2024D03J0008, and Guangzhou Major Talent Project under Grant 2023ZT10X009.

%% file: sec/X_suppl.tex
\clearpage
\setcounter{page}{1}
\maketitlesupplementary

\begin{figure*}
  \centering
  \includegraphics[width=1.0\linewidth]{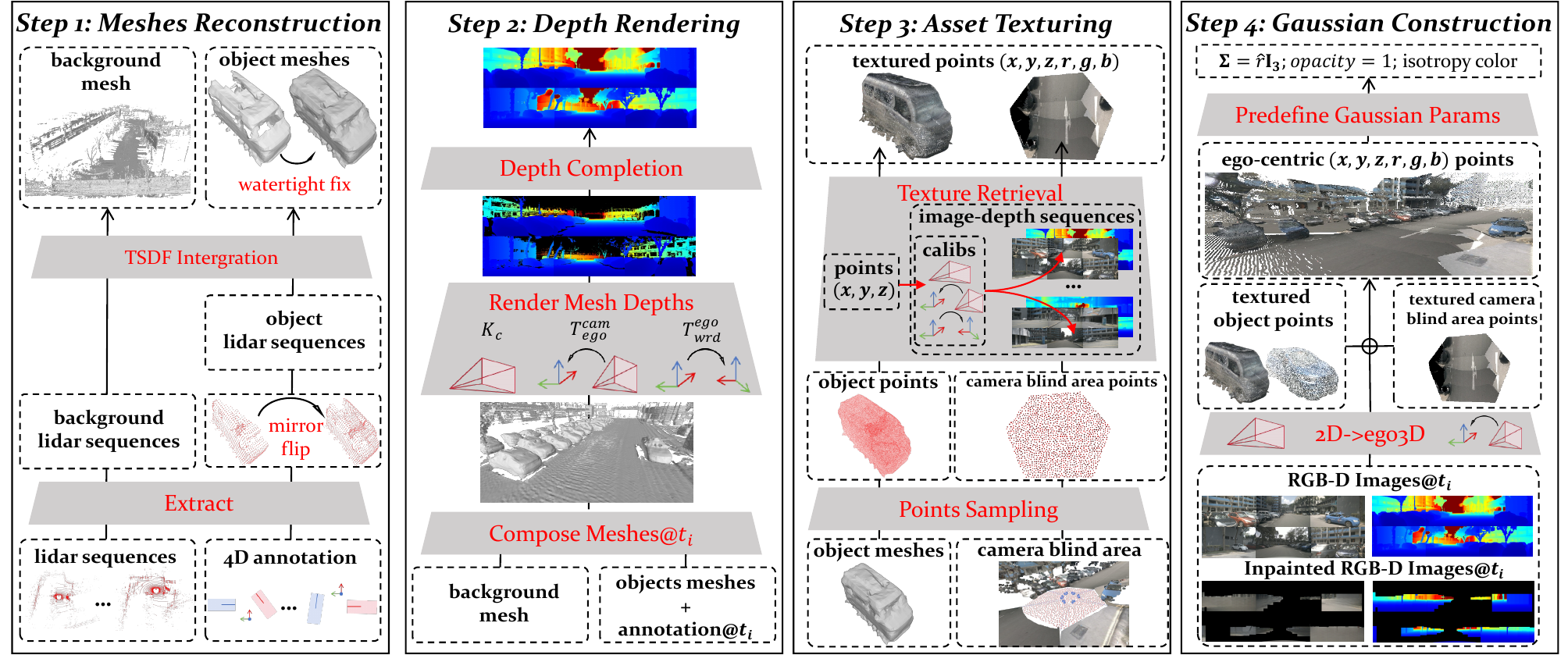}
  \caption{
  The detailed ego-centric Gaussian construction pipeline.
  } 
  \label{fig:training-free-gaussian-contruction-full}
\end{figure*}

\section{Dataset Details}
\label{sec:supp_dataset_details}
The camera configuration details of Nuscene \cite{caesar2020nuscenes}, Lyft \cite{3d-object-detection-for-autonomous-vehicles}, and Waymo \cite{sun2020scalability} are shown in Tab. \ref{table:dataset_infos}. The three dataset have distinct discrepancies in camera configuration, including intrinsics, extrinsics, and array layouts. Given camera intrinsics $(f_u, f_v, c_u, c_v)$ and image size $(w, h)$, the horizontal filed-of-view(FoV) $FoV_{hori}$ and vertical FoV $FoV_{vert}$ are calculated as follows:

\begin{equation}
    \begin{aligned}
        FoV_{hori} = 2\arctan(\frac{c_u}{f_u}) \\
        FoV_{vert} = 2\arctan(\frac{h-c_v}{f_v})
    \end{aligned}
    \label{eq:fov_calc}
\end{equation}

To be noted that, for the three datasets, the ego origin are located at the center of the rear axle projected to the ground. 

\section{Implementation Details}
We show more implementation details about prior maps, data augmentation, model training, and ego-centric Gaussians construction in this section.

\subsection{Prior Maps}
For the prior maps, we normalize them by a constant factor to ensure numerical stability. Specifically, focal length, ground depth, and ground gradient are divided by 500, 25, and 2 respectively. 

\subsection{Data augmentation}
For the data augmentation, denote the raw focal length, camera translation, and camera rotation as $f,t_x,t_y,t_z,r_x,r_y,r_z$, the augmentation policy for each item are $f\sim \mathcal{U}(f*0.7, f*1.4)$, $t_x \sim \mathcal{U} (t_x-0.2,t_x+0.2)$, $t_y \sim \mathcal{U}(t_y-0.2,t_y+0.2)$, $t_z \sim \mathcal{U}\in(1.5, 2.2)$, $r_x \sim \mathcal{U}\in(-2, 2)$, $r_y \sim \mathcal{U}\in(-2, 2)$, $r_z \sim \mathcal{U}\in(r_z-20, r_z+20)$. We randomly select raw image or ego-centric Gaussians for training with probability 0.5.

\subsection{Model Training}
For Waymo, consider the high annotation rate (10$Hz$ vs. 2$Hz$ for Nuscenes) , we select each 4$^{th}$ frame in all sequences for training. We will not downsample data during testing. During training and testing, we apply timestamp alignment for each camera in Nuscenes and Lyft follow by Waymo \cite{sun2020scalability}. All experiments on main text are trained for 24 epochs. 

\input{tables/dataset_infos}

\input{tables/supp_mix_training}

\input{tables/supp_input_adjust_cmp}

\subsection{Ego-centric Gaussian Construction}
The detailed ego-centric Gaussian construction pipeline is shown in Fig. \ref{fig:training-free-gaussian-contruction-full}.

Specifically, the training-free 3DGS can divided into four steps as shown. First, for each data sequence, we use 4D annotations to decompose raw LiDAR sequences into background LiDAR sequences and object LiDAR sequences. Specifically, for each object LiDAR slice, we apply mirror flip to compensate the occluded part. Next, we use TSDF integration \cite{vizzo2022vdbfusion} to reconstruct background mesh and object meshes from background LiDAR sequences and object LiDAR sequences. Moreover, we will fix the object meshes to watertight surface.

Second, for each annotated timestamp in the sequence, we first transform objects meshes from local system to global system according to the annotation. We then compose background mesh and objects meshes. Next, we use camera intrinsic and extrinsic to render the composed meshes to multi-view depth \cite{ravi2020accelerating}. These mesh-rendered depths have advantage that its metric is naturally precise and it can ensure multi-view consistent. These merits are well-suit for our data augmentation. Finally, considerate that some hole will existed in the mesh depths, we apply depth completion \cite{wang2024scale} to get dense depths.

Third, we reconstruct some auxiliary assets including texture objects points model and under-camera points model. These assets can help complete the unseen part beyond raw camera imaging. For example, the unseen under-camera area will affect the novel-view synthesis when we adjust the camera mounting height lower, the unseen object area will affect the NVS when we adjust the camera mounting height higher. To handle this, we first down-sample points from object meshes to get object points model, and we will sample points under camera blind area which can be determined by the initial ground depth mentioned at Sec.3.2 . After constructing points model, we apply texture retrieval for each point by warping them to different images and depths of the sequences. Once the projected depth and the corresponding image depth match under a error threshold, we will allocate the corresponding color to the point. Finally, we can get the textured points model for next step.

Fourth, we construct the ego-view Gaussians for each data frame at each annotated timestamp. First, we use camera intrinsic and extrinsic to project RGB-D images to ego space to get $(r,g,b,x,y,z)$ point cloud. We use Zits++ \cite{cao2023zits++} to inpaint the foreground RGB, and we use ground and background depth to inpaint the foreground depth\cite{kuang2026object}. Then, we concatenate the textured object points and under-camera points to the scene point cloud. Finally, we set covariance matrix of gaussian as $\mathbf{\Sigma}=\hat{r}\mathbf{I}_3$, which means that each Gaussians have no rotation and have a fixed predefined radius. The opacity of each gaussian is set to 1. We set isotropy color for each Gaussians which means that we use the raw $(r,g,b)$ color, the center of Gaussians are determined by the $(x,y,z)$ of point clouds. In other words, our scheme can be seen as a point-rendering scheme but we transform point cloud into Gaussians representation to take advantage of the fast rendering speed of 3DGS.

For the Gaussians radius $\hat{r}$, we set foreground objects Gaussians radius as 0.0025. As for the background, consider that small radius for ground Gaussians will cause hole in image after novel view synthesis, we linearly decrease the $\hat{r}$ from ground plane to sky according to their $z$. Specifically, $\hat{r}$ of background Gaussians with $z$ at range from $(0m, 10m)$ will be linearly mapped to range from $(0.02, 0.001)$. 

\input{tables/supp_nvs_trans}

\input{tables/supp_ground_cc}

\section{More Experiment Results}
\subsection{Multi-Dataset Joint Training}
To evaluate the scaling-up ability of CoIn3D, we conduct joint training using Nuscenes, Waymo, and Lyft. Tab. \ref{table:supp_mix_training} shows that, simply mixing multiple datasets for joint training cannot scale up the performance, because there is a large camera configuration gap between different datasets. CoIn3D can bring significant improvement by bridging the configuration gap. 

\subsection{More Baseline Comparisions}
Tab. \ref{table:supp_input_adjust_cmp} shows a comparison of two input-adjustment methods: Cross-dataset sensor alignment \cite{zheng2023cross} and UniDrive \cite{li2024unidrive}. The results show that they struggle to handle large configuration gaps, especially in N $\rightarrow$ W, because warping inputs destroys the scene geometry structure.

\subsection{Domain-controlled Cfg Generalization}
To control the dataset domain and strengthen the claim (configuration gap) of CoIn3D. We re-render Nuscenes into Nuscenes configuration (N-in-N) and Waymo configuration (N-in-W) using our training-free 3D Gaussians construction pipeline. In this way, we can alleviate the semantic gap and other configuration-irrelevant gaps. We then use N-in-N and N-in-W for training, respectively, and evaluate on N-in-W view. Tab. \ref{table:supp_nvs_trans} shows that CoIn3D almost bridges the NDS* gap to unseen W-Cfg oracle.

\subsection{Ground Corner Cases Evaluation}
To evaluate the robustness of CoIn3D on corner cases under ground plane assumption, we evaluate CoIn3D on 6 sequences with the largest slope (16-22m elevation motion, selected using the 3-$\sigma$ principle) in W-val. Tab. \ref{table:supp_ground_cc} shows that CoIn3D still works well in roads with significant slope. Specifically, CoIn3D trained on either N or W perform well on these scene, even better than the overall performance. This indicates that: (1) ground prior embedding is robust because we do not constrain the ground to fixed depth like UniDrive, but embed it in a learnable way; (2) Further per-sequence analysis reveals that sequences with poor performance are characterized by low light and poor object visibility, which affect both N $\rightarrow$ W$^{'}$ and W $\rightarrow$ W$^{'}$, and are configuration-irrelevant.

\subsection{Source Performance}
Tab. \ref{table:src_performance} shows the results on source dataset after applying our framework. Our framework improve the performance on Nuscenes and Lyft. Regarding to Waymo, the performance drop is mainly attributed to our downsampled training strategy. 

\input{tables/src_performance}

\input{tables/supp_ablation_priors}

\subsection{More ablations}
In this section, we show more ablation experiments regrading to priors map and embedding projector. These experiments are also conduct on BEVDepth by training on Nuscenes and testing on Waymo, but trained for 6 epochs.

Tab. \ref{table:supp_ablation_priors} shown more experiments about the spatial priors, all experiments are based on CDA applied. Specifically, ``FSDepth'' denotes using a focal embedding proposed by \cite{wei2024fs}, ``SFM@(InvF)'' denotes using our inverse focal map to modulate feature, ``SFM@(InvF,GD)'' and ``SFM@(InvF,PR)'' denote using ground depth and plucker raymap additionally based on focal-invariant feature obtained. ``Cat@(ALL)'' denotes only concatenate the prior maps without any modulation. ``SFM@(ALL)'' denotes our final framework. 

The experiment yielded several conclusions. First, using inverse focal map to modulate feature is a better solution compare to FSDepth \cite{wei2024fs} in MC3D. Second, based on focal-invariant feature obtained, solely apply ground depth map or Plücker raymap both benefit the generalization, in view of the former offer explicit ground prior and the latter holistically formulate camera configurations. Third, solely concatenate prior maps without any modulation shown insufficient effectiveness, although explicit prior information is provided, the model is hard to understand these prior without designed modulation mechanism.

Tab. \ref{table:supp_ablation_projector} shows more experiments about the projector structure. Specifically, ``conv K'' and ``conv K + conv K'' denote using one $K\times K$ convolution and two consecutive $K\times K$ to project raw priors into spatial embeddings. Results show that, ``conv3'' is the better choice. Compare to ``conv1'', it provide a large receptive field to fuse the neighbor priors, compare to ``conv3+conv3'', the shallow structure ensures that it does not overfit.

\begin{figure*}
  \centering
  \includegraphics[width=1.0\linewidth]{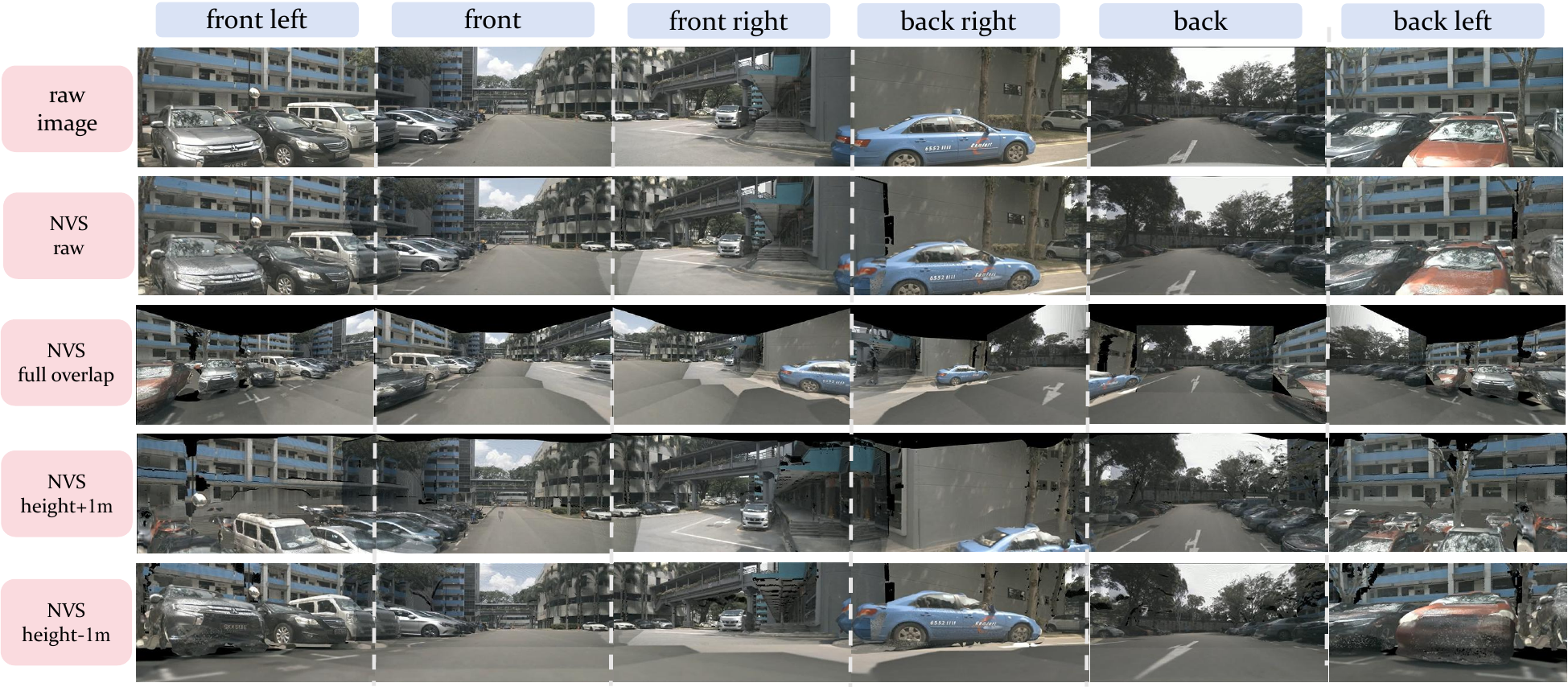}
  \caption{
  Qualitatively visualization of novel view images of Nuscenes.
  } 
  \label{fig:nuscenes_nvs_vis}
\end{figure*}

\begin{figure*}
  \centering
  \includegraphics[width=1.0\linewidth]{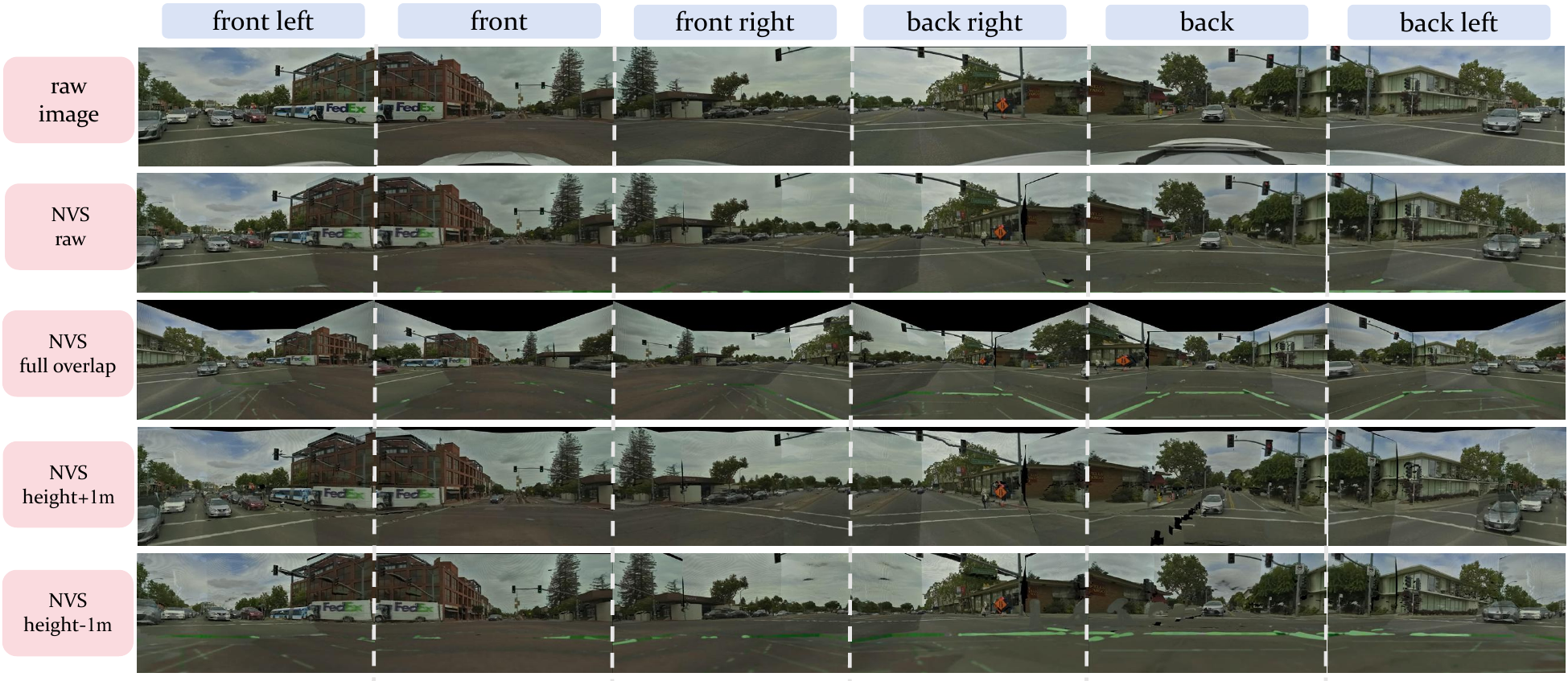}
  \caption{
  Qualitatively visualization of novel view images of Lyft.
  } 
  \label{fig:lyft_nvs_vis}
\end{figure*}

\begin{figure*}
  \centering
  \includegraphics[width=1.0\linewidth]{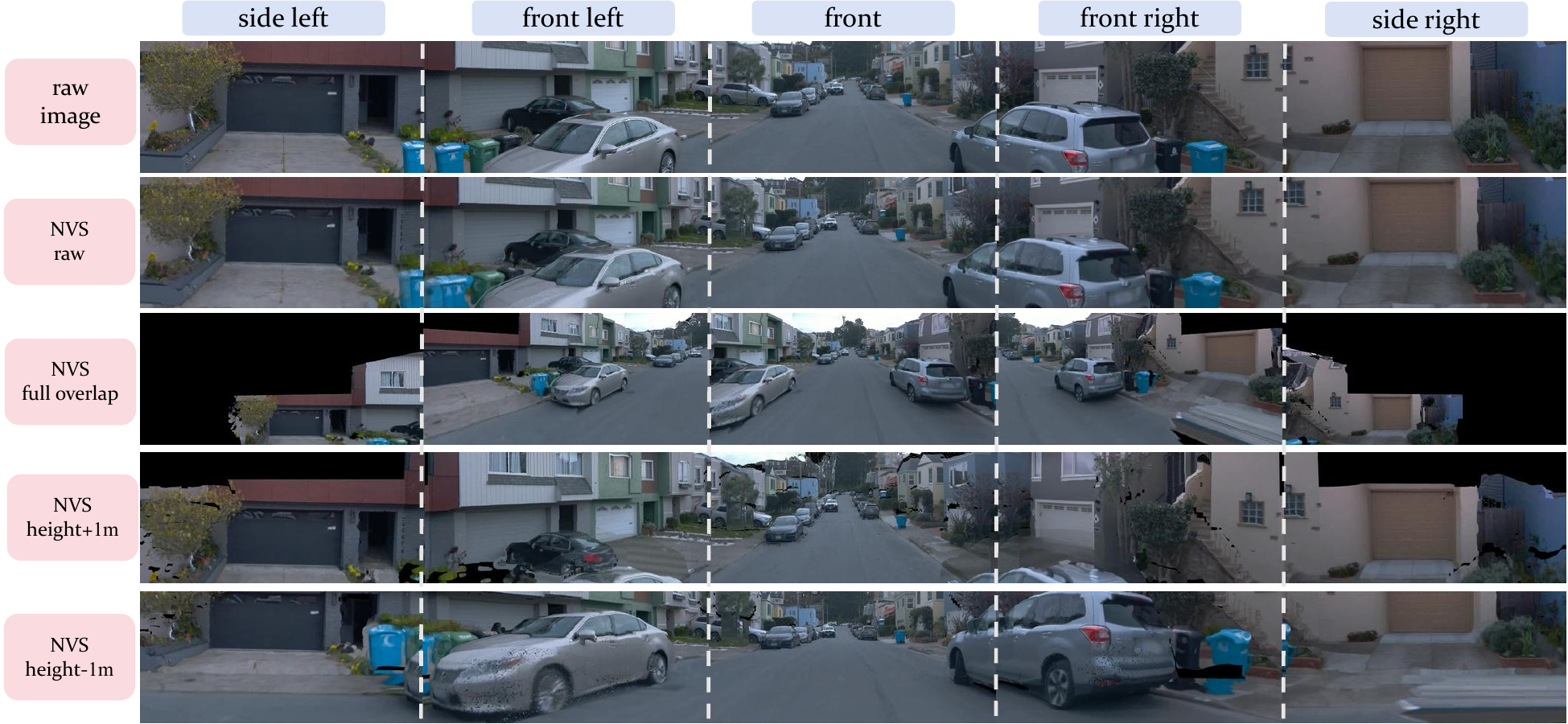}
  \caption{
  Qualitatively visualization of novel view images of Waymo.
  } 
  \label{fig:waymo_nvs_vis}
\end{figure*}

\begin{figure*}
  \centering
  \includegraphics[width=1.0\linewidth]{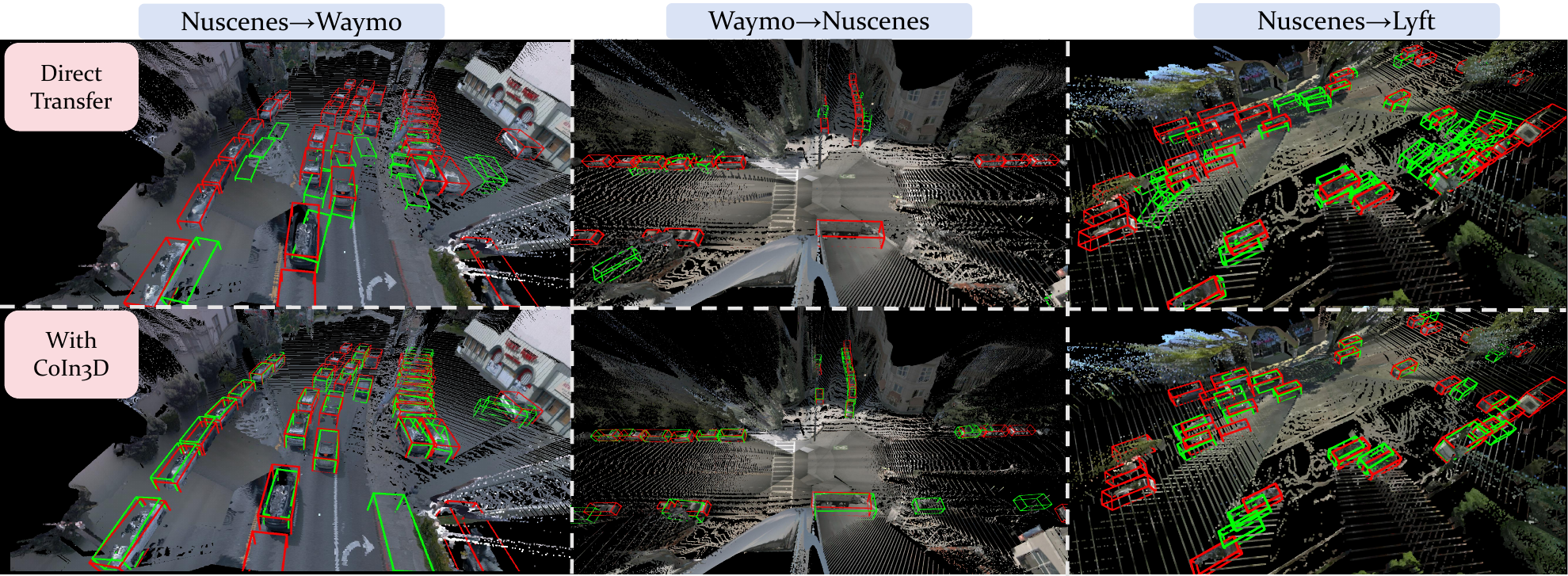}
  \caption{
  Qualitatively visualization of prediction results under 3d textured point cloud. Red boxes denote the ground truth boxes and green boxes denote the predicted boxes. The upper row shows results by directly using raw BEVDepth to inference on target configuration. The bottom row shows results by applying our CoIn3D. 
  } 
  \label{fig:gt_pred_cmp_vis}
\end{figure*}

\input{tables/supp_ablation_projector}

\section{Qualitatively Results}
In this section, we show the qualitatively results about the reconstructed ego-centric textured point clouds and the rendered novel-view images. Moreover, we will also show the detection results with and without using our framework. 

Fig. \ref{fig:nuscenes_nvs_vis}, Fig. \ref{fig:lyft_nvs_vis}, and Fig. \ref{fig:waymo_nvs_vis} show the novel view training images of Nuscenes, Lyft, and Waymo render from the ego-centric Gaussians reconstructed by our training-free Gaussians Construction scheme. We select four setting to compare. ``NVS raw'' denote render image under the raw camera. Under ``NVS full overlap'', we set the horizontal FoV of all cameras as 120°, and we set each two neighbor camera with 60° yaw rotation to construct a full overlap array layout. Under ``NVS height+1m'', we rise the mounting height of each camera by 1m. Under ``NVS height-1m'', we lower the mounting height of each camera by 1m. These figure show that our low-cost training-free ego-centric 3D Gaussians Construction have acceptable photometric reconstruction quality. To be noted that, the reconstruction geometry is naturally precise. 

Fig. \ref{fig:gt_pred_cmp_vis} visualize the 3D detection results on the reconstructed textured point cloud under three cross-configuration testing settings. Fig. \ref{fig:gt_pred_cmp_vis} shows that, use raw base model BEVDepth \cite{li2023bevdepth} to inference on data with new camera configuration results in terrible performance. After applying our CoIn3D, the cross-configuration generalization performance of model can be greatly enhanced. Then model can effectively and precisely detect 3D objects under new camera configuration.

%

%% file: tables/dataset_infos.tex
\begin{table*}[t]
\caption{The details of camera configurations of Nuscenes, Lyft, and Waymo. Camera flags F, FL, FR, B, BL, BR, SL, SR represent front, front left, front right, back, back left, back right, side left, and side right, respectively. Nuscenes and Lyft possess F, FL, FR, B, BL, and BR. Waymo possesses F, FL, FR, SL, and SR. Consider that camera installed pitch ($t_y$) and roll ($t_x$) angle is generally set to 0, we only show the yaw angle ($t_z$).}
\centering

\begin{tabular}{@{}cc|c|c|c|c@{}}
\toprule
\multicolumn{2}{c|}{Datasets}                                              & Nuscenes                    & Lyft(fleet 1)                & Lyft(fleet 2)                & Waymo                 \\ \midrule
\multicolumn{2}{c|}{\multirow{2}{*}{image resolution$(w,h)$}}                & \multirow{2}{*}{(1600,900)} & \multirow{2}{*}{(1224,1024)} & \multirow{2}{*}{(1920,1080)} & (1920,1280)(F,FL,FR)  \\
\multicolumn{2}{c|}{}                                                      &                             &                              &                              & (1920,886)(SL,SR)     \\ \midrule
\multicolumn{2}{c|}{\multirow{2}{*}{focal length}}                         & 1250(F,FL,FR,BL,BR)         & \multirow{2}{*}{880}         & \multirow{2}{*}{1100}        & \multirow{2}{*}{2050} \\
\multicolumn{2}{c|}{}                                                      & 800(B)                      &                              &                              &                       \\ \midrule
\multicolumn{2}{c|}{\multirow{2}{*}{horizontal FoV (°)}}                        & 65(F,FL,FR,BL,BR)           & \multirow{2}{*}{70}          & \multirow{2}{*}{80}          & \multirow{2}{*}{50}   \\
\multicolumn{2}{c|}{}                                                      & 90(B)                       &                              &                              &                       \\ \midrule
\multicolumn{2}{c|}{\multirow{2}{*}{vertical FoV (°)}}                         & 40(F,FL,FR,BL,BR)           & \multirow{2}{*}{60}          & \multirow{2}{*}{50}          & \multirow{2}{*}{35}   \\
\multicolumn{2}{c|}{}                                                      & 60(B)                       &                              &                              &                       \\ \midrule
\multicolumn{2}{c|}{cam numbers}                                           & 6                           & 6                            & 6                            & 5                     \\ \midrule
\multicolumn{1}{c|}{\multirow{6}{*}{\shortstack{camera translation \\$(t_x,t_y,t_z)$ (m)}}} & F     & 1.7, 0, 1.5                   & 1.5, 0, 1.7                    & 1.5, 0, 1.65                   & 1.55, 0, 2.1            \\
\multicolumn{1}{c|}{}                                              & FL    & 1.55, 0.5, 1.5                & 1.3, 0.3, 1.7                  & 1.3, 0.3, 1.65                 & 1.5, 0.1, 2.1           \\
\multicolumn{1}{c|}{}                                              & FR    & 1.55, -0.5, 1.5               & 1.3, -0.3, 1.7                 & 1.3, -0.3, 1.65                & 1.5, -0.1, 2.1          \\
\multicolumn{1}{c|}{}                                              & B     & 0, 0, 1.5                     & 0.8, 0, 1.65                   & 0.8, 0, 1.65                   &                       \\
\multicolumn{1}{c|}{}                                              & BL/SL & 1.0, 0.5, 1.55                & 1.0, 0.3, 1.65                 & 1.0, 0.3, 1.65                 & 1.4, 0.1, 2.1           \\
\multicolumn{1}{c|}{}                                              & BR/SR & 1.0, -0.5, 1.55               & 1.0, -0.3, 1.65                & 1.0, -0.3, 1.65                & 1.4, -0.1, 2.1          \\ \midrule
\multicolumn{1}{c|}{\multirow{6}{*}{cam rotation $(r_z)$ (°)}}             & F     & 0                           & 0                            & 0                            & 0                     \\
\multicolumn{1}{c|}{}                                              & FL    & 55                          & 60                           & 60                           & 45                    \\
\multicolumn{1}{c|}{}                                              & FR    & -55                         & -60                          & -60                          & -45                   \\
\multicolumn{1}{c|}{}                                              & B     & 180                         & 180                          & 180                          &                       \\
\multicolumn{1}{c|}{}                                              & BL/SL    & 110                         & 120                          & 120                          & 90                    \\
\multicolumn{1}{c|}{}                                              & BR/SR    & -110                        & -120                         & -120                         & -90                   \\ \bottomrule
\end{tabular}

\label{table:dataset_infos}
\end{table*}

%% file: tables/supp_mix_training.tex
\begin{table*}[t]
\caption{Results of Nuscenes+Waymo+Lyft joint training (Mix), reported on 3 classes(car, pedestrian, two-wheel object). ScaleBEV \cite{lu2024scaling} are reported on BEVDet, others are reported on BEVDepth. Oracle means training on the single source dataset.}
\centering

\begin{tabular}{@{}c|c|cc|cc|cc|cc@{}}
\toprule
\multirow{2}{*}{Setting}                    & \multirow{2}{*}{Method} & \multicolumn{2}{c|}{All}        & \multicolumn{2}{c|}{Car}        & \multicolumn{2}{c|}{Pedestrian} & \multicolumn{2}{c}{Two-wheel}   \\
                                            &                         & mAP            & NDS*           & mAP            & NDS*           & mAP            & NDS*           & mAP            & NDS*           \\ \midrule
\multirow{4}{*}{Mix $\rightarrow$ Nuscenes} & Oracle                  & 0.255          & 0.335          & 0.385          & 0.508          & 0.205          & 0.247          & 0.175          & 0.249          \\
                                            & Direct Mix              & 0.267          & 0.350          & 0.411          & 0.537          & 0.213          & 0.258          & 0.176          & 0.255          \\
                                            & ScaleBEV                & 0.293          & 0.358          & 0.447          & 0.540          & 0.246          & 0.275          & 0.185          & 0.258          \\
                                            & \textbf{CoIn3D}           & \textbf{0.358} & \textbf{0.441} & \textbf{0.542} & \textbf{0.693} & \textbf{0.321} & \textbf{0.360} & \textbf{0.211} & \textbf{0.270} \\ \midrule
\multirow{4}{*}{Mix $\rightarrow$ Lyft}     & Oracle                  & 0.290          & 0.390          & 0.408          & 0.530          & 0.114          & 0.199          & 0.349          & 0.440          \\
                                            & Direct Mix              & 0.312          & 0.406          & 0.473          & 0.581          & 0.126          & 0.201          & 0.338          & 0.436          \\
                                            & ScaleBEV                & 0.333          & 0.422          & 0.505          & 0.603          & 0.140          & 0.212          & 0.355          & 0.453          \\
                                            & \textbf{CoIn3D}           & \textbf{0.455} & \textbf{0.532} & \textbf{0.644} & \textbf{0.718} & \textbf{0.236} & \textbf{0.323} & \textbf{0.486} & \textbf{0.556} \\ \midrule
\multirow{4}{*}{Mix $\rightarrow$ Waymo}    & Oracle                  & 0.298          & 0.376          & 0.426          & 0.536          & 0.334          & 0.328          & 0.135          & 0.265          \\
                                            & Direct Mix              & 0.306          & 0.373          & 0.432          & 0.536          & 0.342          & 0.332          & 0.144          & 0.250          \\
                                            & ScaleBEV                & 0.326          & 0.384          & 0.466          & 0.549          & 0.355          & 0.341          & 0.157          & 0.261          \\
                                            & \textbf{CoIn3D}           & \textbf{0.381} & \textbf{0.463} & \textbf{0.559} & \textbf{0.647} & \textbf{0.392} & \textbf{0.363} & \textbf{0.192} & \textbf{0.381} \\ \bottomrule
\end{tabular}

\label{table:supp_mix_training}
\end{table*}

%% file: tables/supp_input_adjust_cmp.tex
\begin{table}[]
\caption{Comparison of CoIn3D and two input-adjustment methods: Cross-dataset sensor alignment (CDSA) \cite{zheng2023cross} and Unidrive \cite{li2024unidrive}.}
\centering

\begin{tabular}{@{}c|cc@{}}
\toprule
Setting                           & mAP            & \textbf{NDS*}  \\ \midrule
CDSA N $\rightarrow$ L            & 0.224          & 0.372          \\
UniDrive N $\rightarrow$ L        & 0.264          & 0.440          \\
\textbf{CoIn3D N $\rightarrow$ L} & \textbf{0.375} & \textbf{0.534} \\ \midrule
CDSA N $\rightarrow$ W            & 0.116          & 0.259          \\
UniDrive N $\rightarrow$ W        & 0.144          & 0.298          \\
\textbf{CoIn3D N $\rightarrow$ W} & \textbf{0.384} & \textbf{0.513} \\ \bottomrule
\end{tabular}

\label{table:supp_input_adjust_cmp}
\end{table}

%% file: tables/supp_nvs_trans.tex
\begin{table}[]
\caption{Results of Nuscenes-val re-rendered in Waymo Cfg.}
\centering

\begin{tabular}{@{}c|cc@{}}
\toprule
Setting         & mAP            & \textbf{NDS*}  \\ \midrule
Oracle          & 0.436          & 0.554          \\
DT              & 0.126          & 0.342          \\
\textbf{CoIn3D} & \textbf{0.401} & \textbf{0.524} \\ \bottomrule
\end{tabular}

\label{table:supp_nvs_trans}
\end{table}

%% file: tables/supp_ground_cc.tex
\begin{table}[]
\caption{Results of sequences with significant road slope in W-val (W$^{'}$). Results are reported on BEVDepth with CoIn3D.}
\centering

\begin{tabular}{@{}c|cc@{}}
\toprule
Setting         & mAP            & \textbf{NDS*}  \\ \midrule
N $\rightarrow$ W$^{'}$         & 0.516          & 0.602          \\
W $\rightarrow$ W$^{'}$              & 0.638          & 0.702          \\
\bottomrule
\end{tabular}

\label{table:supp_ground_cc}
\end{table}

%% file: tables/src_performance.tex
\begin{table}[]
\caption{Source dataset performance comparison.}
\centering

\begin{tabular}{@{}cc|cc@{}}
\toprule
Dataset                   & Method   & mAP$\uparrow$   & NDS$^*$$\uparrow$   \\ \midrule
\multirow{3}{*}{Nuscenes} & BEVDepth \cite{li2023bevdepth} & 0.475 & 0.587 \\
                          & UDGA-BEV \cite{chang2024unified} & 0.497 & 0.603 \\
                          & Ours     & \textbf{0.529} & \textbf{0.630} \\ \midrule
\multirow{3}{*}{Lyft}     & BEVDepth & 0.602 & 0.684 \\
                          & UDGA-BEV & \textbf{0.630} & \textbf{0.702} \\
                          & Ours     & 0.624 & 0.699 \\ \midrule
\multirow{3}{*}{Waymo}    & BEVDepth & 0.552 & 0.649 \\
                          & UDGA-BEV & \textbf{0.547} & \textbf{0.656} \\
                          & Ours     & 0.538 & 0.630 \\ \bottomrule
\end{tabular}

\label{table:src_performance}
\end{table}

%% file: tables/supp_ablation_priors.tex
\begin{table}[]
\caption{More ablation results regrading to priors.}
\centering

\begin{tabular}{@{}c|cc@{}}
\toprule
Method             & mAP$\uparrow$            & NDS$^*$$\uparrow$            \\ \midrule
FSDepth \cite{wei2024fs}            & 0.223          & 0.374          \\
SFM@(InvF)         & 0.260          & 0.404          \\
SFM@(InvF,GD)      & 0.274          & 0.425          \\
SFM@(InvF,PR)      & 0.272          & 0.429          \\
Cat@(ALL)          & 0.247          & 0.391          \\
\textbf{SFM@(ALL)} & \textbf{0.320} & \textbf{0.463} \\ \bottomrule
\end{tabular}

\label{table:supp_ablation_priors}
\end{table}

%% file: tables/supp_ablation_projector.tex
\begin{table}[]
\caption{More ablation results regrading to embedding projectors.}
\centering

\begin{tabular}{@{}c|cc@{}}
\toprule
Method      & mAP$\uparrow$   & NDS$^*$$\uparrow$   \\ \midrule
conv1       & 0.282 & 0.430 \\
conv1+conv1 & 0.307 & 0.454 \\
conv3+conv3 & 0.297 & 0.448 \\
conv3       & \textbf{0.320} & \textbf{0.463} \\ \bottomrule
\end{tabular}

\label{table:supp_ablation_projector}
\end{table}